\documentclass[10pt,twocolumn,letterpaper]{article}

\usepackage{cvpr}
\usepackage{times}
\usepackage{epsfig}
\usepackage{graphicx}
\usepackage{amsmath}
\usepackage{amssymb}

\usepackage{multirow}
\usepackage{threeparttable}
\usepackage[table,dvipsnames]{xcolor}
\usepackage{breqn}
\usepackage{gensymb}
\usepackage{tabu}
\usepackage{arydshln}
\usepackage[super]{nth}
\usepackage{accsupp}
\usepackage[pagebackref=true,breaklinks=true,letterpaper=true,colorlinks,bookmarks=false]{hyperref}
\cvprfinalcopy %

\def\cvprPaperID{5655} %

\usepackage{textcomp}
\begin{document}

\title{CAM-Convs: Camera-Aware Multi-Scale Convolutions for Single-View Depth}
\author{Jose M. Facil\textsuperscript{1}\quad Benjamin Ummenhofer\textsuperscript{2,3}\quad Huizhong Zhou\textsuperscript{2}\\
{\tt\small jmfacil@unizar.es}\quad {\tt\small benjamin.ummenhofer@intel.com} \quad {\tt\small zhouh@cs.uni-freiburg.de}\\
Luis Montesano\textsuperscript{1,4}\quad Thomas Brox\textsuperscript{*,2} \quad Javier Civera\textsuperscript{*,1}\\
{\tt\small montesano@unizar.es} \quad {\tt\small brox@cs.uni-freiburg.de} \quad {\tt\small jcivera@unizar.es}\\
\textsuperscript{1} University of Zaragoza\quad\textsuperscript{2} University of Freiburg\quad\textsuperscript{3} Intel Labs\quad\textsuperscript{4} Bitbrain\\
}

\maketitle
\renewcommand*{\thefootnote}{\fnsymbol{footnote}}
\footnotetext[1]{Equal contribution}
\let\thefootnote\relax\footnotetext[2]{Find our \textbf{code} in {\scriptsize 
\href{http://webdiis.unizar.es/~jmfacil/camconvs}{\tt webdiis.unizar.es/{\texttildelow}jmfacil/camconvs}}}
\newcommand{\preCua}{LR-1}
\newcommand{\preTre}{MR-1}
\newcommand{\preDos}{MR-2}
\newcommand{\preUno}{HR-1}
\newcommand{\preOri}{HR-2}

\newcommand*{\mrowstyle}{}

\newcommand*{\rowstyle}[1]{%
  \gdef\@rowstyle{#1}%
  \mrowstyle\ignorespaces%
}
\definecolor{orange(webcolor)}{rgb}{1.0, 0.65, 0.0}
\definecolor{orange-red}{rgb}{1.0, 0.27, 0.0}
\definecolor{orange(colorwheel)}{rgb}{1.0, 0.5, 0.0}
\definecolor{internationalorange}{rgb}{1.0, 0.31, 0.0}
\definecolor{iris}{rgb}{0.35, 0.31, 0.81}
\colorlet{baseline1}{internationalorange}
\colorlet{baseline2}{iris}
\newcolumntype{=}{%
  >{\gdef\@rowstyle{}}%
}

\newcolumntype{+}{%
  >{\@rowstyle}%
}
\newcommand{\triast}{\bigskip\par\noindent\parbox{\linewidth}{\centering\large{*}\\[-4pt]{*}\hskip 0.75em{*}}\bigskip\par}%

\newcommand{\smallFocal}{72}
\newcommand{\bigFocal}{128}
\newcommand{\extraFocal}{64}
\newcommand{\normFocal}{100}
\newcommand{\extraIm}{\mbox{$320\times320$}}
\newcommand{\smallIm}{\mbox{$128\times96$}}
\newcommand{\bigIm}{\mbox{$256\times192$}}
\newcommand{\bigImt}{\mbox{$192\times256$}}
\newcommand{\sqIm}{\mbox{$224\times224$}}
\newcommand{\scannetSize}{\mbox{$256\times192$}}
\newcommand{\kittiSize}{\mbox{$384\times128$}}

\newcommand{\smallFocalR}{$f_{72}$}
\newcommand{\bigFocalR}{$f_{128}$}
\newcommand{\extraFocalR}{$f_{64}$}
\newcommand{\normFocalR}{$f_{n}$}
\newcommand{\scannetSizeR}{$s_S$}
\newcommand{\kittiSizeR}{$s_K$}
\newcommand{\extraImR}{$s_5$}
\newcommand{\smallImR}{$s_4$}
\newcommand{\bigImR}{$s_1$}
\newcommand{\bigImtR}{$s_2$}
\newcommand{\sqImR}{$s_3$}

\newcommand{\thu}{$\delta<1.25^1$}
\newcommand{\thd}{$\delta<1.25^2$}
\newcommand{\tht}{$\delta<1.25^3$}

\newcommand{\shortConvs}{CAM-Convs}
\newcommand{\longConvs}{Camera-Aware Multi-scale Convolutions}
\newcommand{\withBoxNet}{CAM-Convs}

\newcommand{\oneBsize}{\mbox{$320\times320$}}
\newcommand{\twoBsize}{\mbox{$256\times256$}}
\newcommand{\threeBsize}{\mbox{$224\times224$}}

\newcommand{\stb}{\textit{smallest the best}}
\newcommand{\btb}{\textit{biggest the best}}
\definecolor{magenta}{rgb}{1.0, 0.0, 1.0}
\definecolor{selectiveyellow}{rgb}{1.0, 0.73, 0.0}
\definecolor{darkpastelgreen}{rgb}{0.01, 0.75, 0.24}

\begin{abstract}
Single-view depth estimation suffers from the problem that a network trained on images from one camera does not generalize to images taken with a different camera model. Thus, changing the camera model requires collecting an entirely new training dataset. In this work, we propose a new type of convolution that can take the camera parameters into account, thus allowing neural networks to learn calibration-aware patterns. Experiments confirm that this improves the generalization capabilities of depth prediction networks considerably, and clearly outperforms the state of the art when the train and test images are acquired with different cameras.

\end{abstract}
\vspace{-2ex}
\section{Introduction}

Recovering 3D information from 2D images is one of the fundamental problems in computer vision that, due to recent advances and applications, is receiving nowadays a renewed attention. Among others, there has been recent relevant results on problems such as 6D object pose detection \cite{kehl2017ssd,shrivastava2017learning,sundermeyer2018implicit}, 3D model reconstruction \cite{fan2017point,tatarchenko2017octree}, depth estimation from single \cite{laina2016deeper,fu2018deep,li2018deep} and multiple views \cite{ummenhofer2017demon,huang2018deepmvs}, 6D camera pose recovery \cite{kendall2015posenet,kendall2017geometric} or camera tracking and mapping \cite{zhou2018deeptam,bloesch2018codeslam,tang2018ba,tateno2017cnn}. 
While traditional multi-view methods (e.g., \cite{schonberger2016structure}) are mostly based on geometry and optimization and, thus, are largely independent of the data, these recent deep learning approaches depend on training data that demonstrates the mapping from images to depth. 

\begin{figure}[t!]
\centering
\includegraphics[width=1.\linewidth]{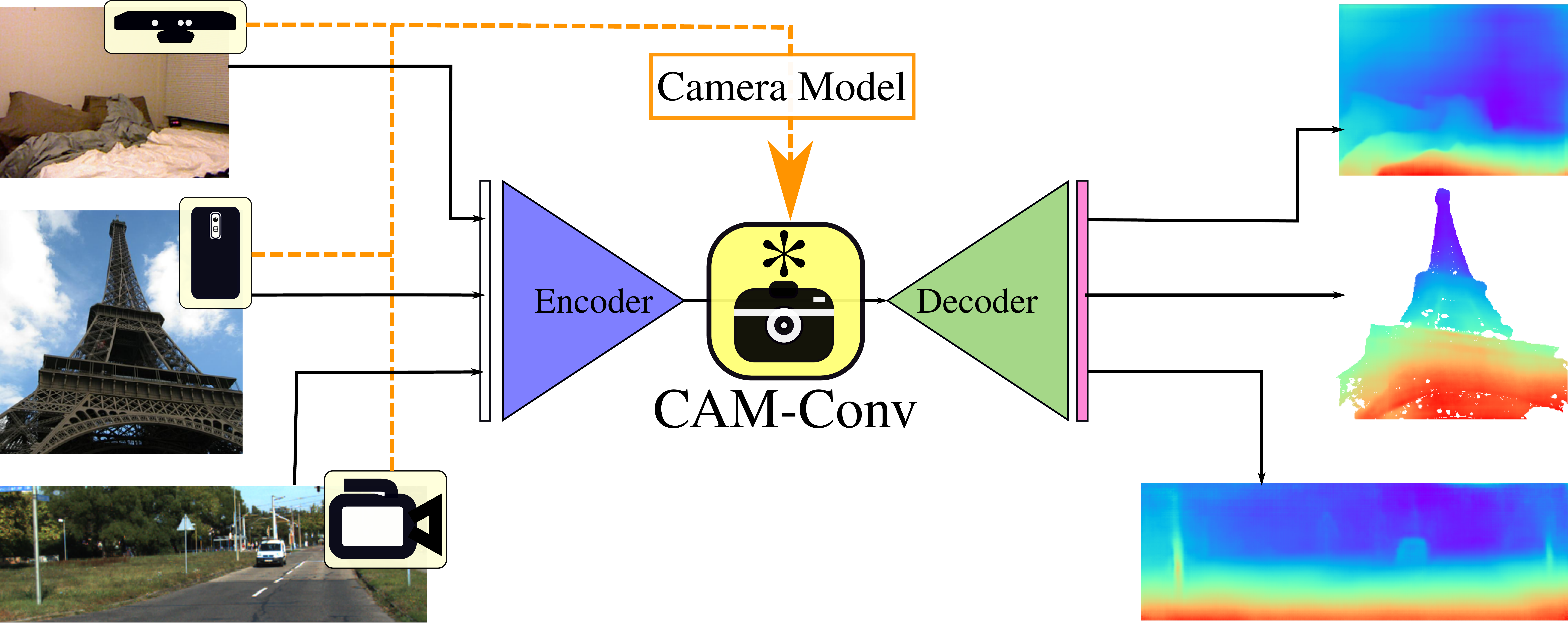}
\caption{\label{fig:teaser} \shortConvs{} allows efficient specialization of a camera-generic network for various camera models by feeding camera-specific parameters into the network.}
\vspace{-3mm}
\end{figure}

The common strategy to collect such data is by using an RGBD sensor, like the Kinect camera, which conveniently provides both the RGB image and what can be considered ground truth depth. It is implicitly assumed that training on this type of data will generalize to other RGB sensors that do not provide depth. However, the evaluation of recent learning-based methods relies largely on public benchmarks where images have been recorded with the same RGBD camera as the training data. Thus, evaluation on these benchmarks does not reveal whether a depth estimation method generalizes to RGB images from another camera. 

Overfitting to a benchmark is a common problem in computer vision research. Other works \cite{torralba2011unbiased} have shown that datasets may have strong biases that make researchers over-confident regarding the performance of their method. In particular, train-test divisions of the same kind of data are not enough to prove generalization. In this work we show that, indeed, state-of-the-art single-view depth prediction networks do not generalize when the camera parameters of the test images are different from the training ones.

Moreover, we show that for single-view depth prediction the problem of missing generalization to images from different cameras is even more severe: it cannot be solved by training on images from a diverse set of cameras with different parameters. For present methods to adapt to a different camera model, they require changes in the architecture.

We present a deep neural network for single-view depth prediction that, for the first time, addresses the variability on the camera's internal parameters. We show that this allows to use images from different cameras at train and test time without a performance degradation. This is of particular interest, as it enables the exploitation of images from \emph{any} camera for training the data-hungry deep networks. Specifically, within our proposed network, our main contribution is a novel type of convolution, that we name as \shortConvs{} (\longConvs{}), that concatenates the camera internal parameters to the feature maps, and hence allows the network to learn the dependence of the depth from these parameters. Figure~\ref{fig:teaser} shows an illustration of how \shortConvs{} act in the typical encoder-decoder depth estimation pipeline. 
The network can be trained with a mixture of images from different cameras without overfitting to specific intrinsics. We show that the network generalizes also to images from cameras it has not been trained on. A comparison with the state of the art in single-image depth estimation demonstrates that the better generalization properties do not reduce the accuracy of the depth estimates. 
\section{Related Work}

Estimating 3D structure and 6 degrees-of-freedom motion using deep learning has been addressed recently from several angles: Supervised \cite{laina2016deeper} and unsupervised \cite{zhou2017unsupervised}, from single \cite{eigen2015predicting} and multiple views \cite{tang2018ba}, using end-to-end networks \cite{laina2016deeper} and fusing with multi-view geometry \cite{facil2017single}, completing depth maps \cite{zhang2018deep,weerasekera2018just}, and estimating geolocation \cite{weyand2016planet,kendall2015posenet}, relative motion \cite{ummenhofer2017demon}, visual odometry \cite{wang2017deepvo,wang2018end}, and simultaneous localization and mapping (SLAM) \cite{tateno2017cnn,bloesch2018codeslam,zhou2018deeptam}. 

In this work we deal with single-view supervised depth learning, so we will focus our literature review in this case. Among the pioneering work we can reference \cite{hoiem2005automatic}, that similarly to pop-up illustrations, cut and fold a 2D image based on a segmentation into geometric classes and some geometric assumptions. \cite{saxena2009make3d} is another seminal work that, with minimal assumptions on the scene, learned a model based on a MRF. \cite{eigen2014depth} was the first paper that used deep learning for single-view depth prediction, proposing a multi-scale depth network. Its results were improved later by \cite{eigen2015predicting,liu2015deep,laina2016deeper,chakrabarti2016depth,he2018learning}.

Many methods focus on specific datasets which enable to train learning-based methods for specific tasks.
For instance, Eigen and Fergus \cite{eigen2015predicting} extend the multi-scale architecture in \cite{eigen2014depth} to the prediction of surface normals and semantic labels on the NYU dataset \cite{silberman2012indoor}.
Similarly, Wang \etal \cite{wang2015towards} train a network that jointly predicts depth and segmentation on the same dataset.
For depth, Laina \etal \cite{laina2016deeper}, Liu \etal \cite{liu2015deep} and Eigen \etal \cite{eigen2015predicting} show that their methods can be adapted to other datasets like Make3D \cite{saxena2009make3d} or KITTI \cite{geiger2012kitti}.
However, they treat datasets like different tasks and require retraining for each dataset to achieve state-of-the-art performance.

Chen \etal \cite{chen2016single-image}, inspired by \cite{zoran2015learning}, introduce the Depth in the Wild dataset and train a CNN using ordinal relations between point pairs. While the images stem from internet photo collections taken with many different cameras, they do not make use of the camera parameters during training.
Li and Snavely \cite{MegaDepthLi18} use a structure from motion pipeline to extract depth from internet photo collections and use this to train a CNN predicting depth up to a scale factor. Again, information about camera  parameters is not exploited and generalization is solely driven by large diverse datasets. Extrinsic parameters have been considered for other tasks such as stereo estimates \cite{ummenhofer2017demon} or synthesis of view point changes \cite{zhou2016view}. Intrinsic parameters are usually left out in deep learning pipelines, with the exception of He \etal \cite{he2018learning}. They embed focal length information in a fully-connected approach, making it impossible to train and test in different image sizes, while our proposal is flexible and can deal with different image sizes.

In the next section we describe how to explicitly implement the internal camera parameters into the network and thereby improve generalization by \shortConvs{}.

\section{\longConvs{}}
\label{sec:camconvs}
\label{sec:addingcamconv}
\begin{figure}[t!]
\centering
\includegraphics[width=1.\linewidth]{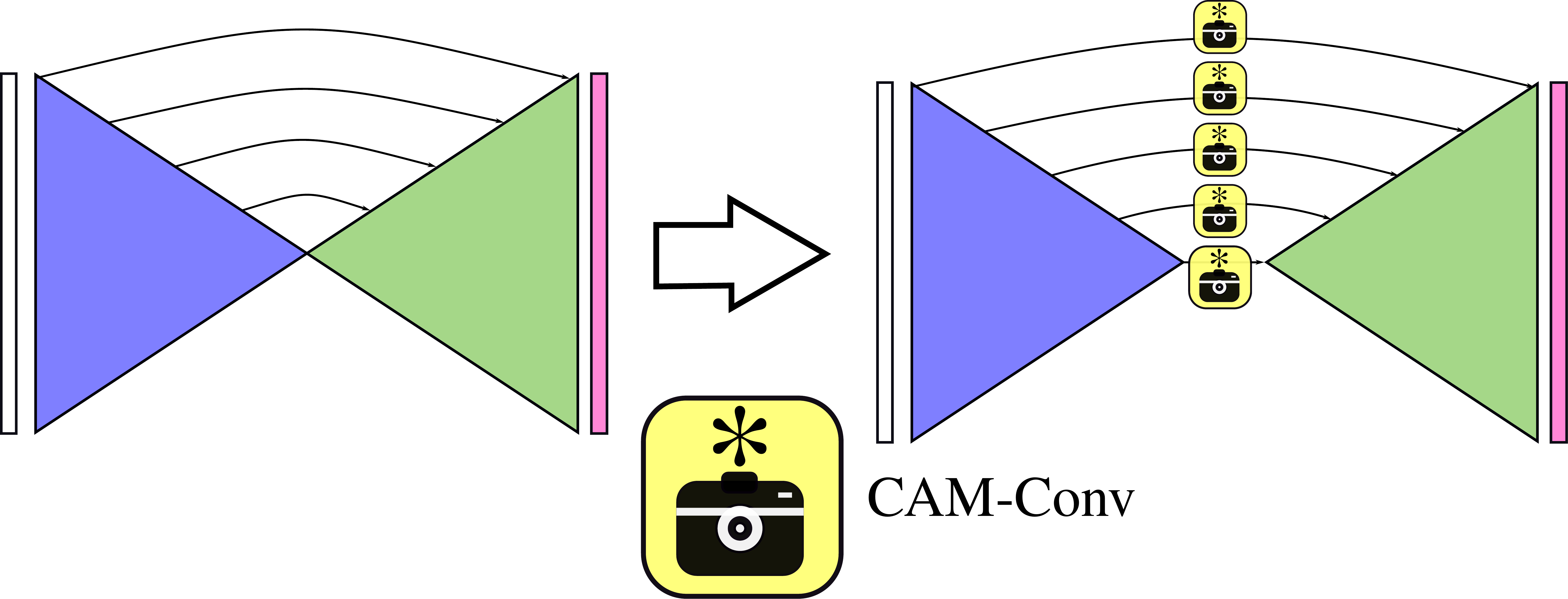}
\caption{\label{fig:incamconvs} Adding \shortConvs{} to an Encoder-Decoder U-Net architecture. }
\vspace{-4mm}
\end{figure}
\shortConvs{} (standing for \longConvs{}), is the variant of the convolution operation that we present in this paper. \shortConvs{} include the camera intrinsics in the convolutions, allowing the network to learn and predict depth patterns that depend on the camera calibration. Specifically, we add \shortConvs{} in the mapping from RGB features to 3D information--\textit{e.g.} depth, normals--, that is, between the encoder and the decoder. As shown in Figure~\ref{fig:incamconvs}, we add them at every level, such that we include \shortConvs{} on every skip-connection too. 
Notice that all the \shortConvs{} are added after the encoder, allowing the use of pretrained models.

\begin{figure}[t!]
\centering
\includegraphics[width=1.\linewidth]{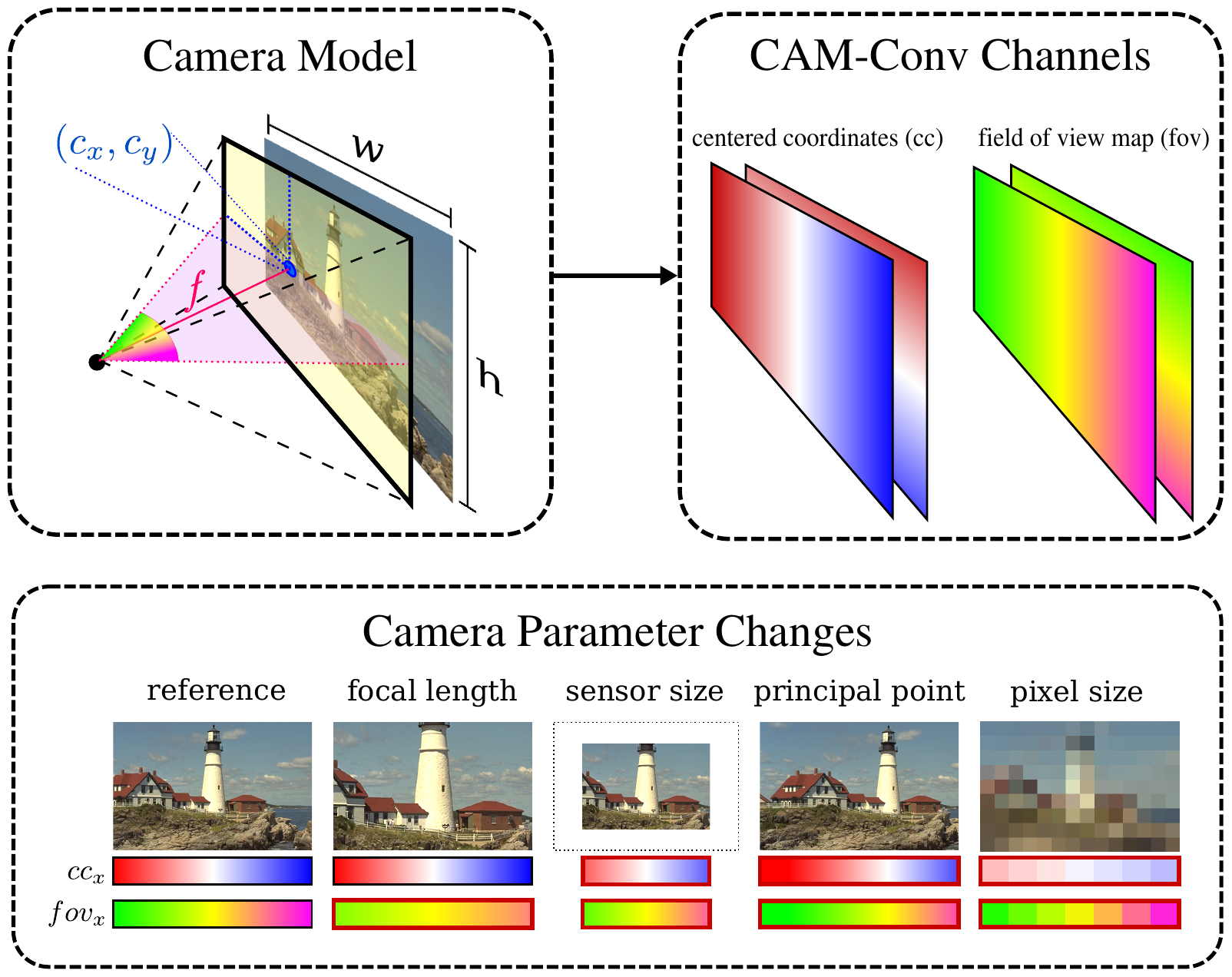}
\caption{\label{fig:camconvs} Overview of the additional channels of our \shortConvs{} (\longConvs{}). We compute Centered Coordinates ($cc$ from {\color{red} red}  to {\color{blue}blue}) and Field of View ($fov$ from {\color{darkpastelgreen}green} to {\color{magenta}pink})  maps. We concatenate these maps with the input features before applying the convolution. Both $cc$ and $fov$ depend on the camera model and are sensitive to camera changes. The Bottom part shows how $cc$ and $fov$ maps change with the camera parameters (a {\color{red} red border} means the map has changed from the original). %
}
\vspace{-3mm}
\end{figure}

 The basics of \shortConvs{} are as follows: We pre-compute pixel-wise coordinates and field-of-view maps and feed them along with the input features to the convolution operation. \shortConvs{} use the idea behind Coord-Convs \cite{liu2018intriguing}, on adding normalized coordinates per pixel, but incorporating information on the camera calibration. An illustrative scheme of how \shortConvs{} extra channels work is shown in Figure~\ref{fig:camconvs}. The different maps included are computed using the camera intrinsic parameters (focal length $f$ and principal point coordinates $(c_x,c_y)$) and the sensor size (width $w$ and height $h$):

\textbf{Centered Coordinates ($cc$):} 
To add the information of the principal point location to the convolutions, we include $cc_x$ and $cc_y$ coordinate channels centered at the principal point--i.e. the principal point has coordinates $(0,0)$. Specifically, the channels are
\begin{equation}
    \label{eq:ccx}
    cc_x=\begin{bmatrix}
           \smash{0 -c_x}\\
           \smash{1 -c_x}\\
           \smash{\vdots} \\
           \smash{w-c_x}
    \end{bmatrix}_{\smash{w\times 1}}\cdot\begin{bmatrix}
           1 \\
           1 \\
           \smash{\vdots} \\
           1
    \end{bmatrix}_{h\times 1}^{\intercal} = \begin{bmatrix} -c_x & {\cdots} & w-c_x \\ {\vdots} & {\ddots} & {\vdots} \\ -c_x & {\cdots} & w-c_x \end{bmatrix}
  \end{equation}
  \begin{equation}
  \label{eq:ccy}
    cc_y=\begin{bmatrix}
           1 \\
           1 \\
           \smash{\vdots} \\
           1
    \end{bmatrix}_{w\times 1}\cdot\begin{bmatrix}
           \smash{0 -c_y}\\
           \smash{1 -c_y}\\
           \smash{\vdots} \\
           \smash{h - c_y}
    \end{bmatrix}_{h\times 1}^{\intercal}=\begin{bmatrix} -c_y & \cdots & -c_y \\ \vdots & \ddots & \vdots \\ h - c_y & \cdots & h - c_y \end{bmatrix}.
\end{equation}

We resize these maps to the input feature size using bilinear interpolation and concatenate them as new input channels. These channels are sensitive to the sensor size and resolution (pixel size) of the camera, as their values depend on it. We assume the sensor size is measured in pixels. In Figure \ref{fig:camconvs} we represent $cc$ with a  color gradient from {\color{red} red} (for negative coordinates) to {\color{blue}blue}  (for positive coordinates), white for 0. Notice in the figure how $cc$ values change when camera sensor size, principal point or pixel size change.

\textbf{Field of View Maps ($fov$):} The horizontal and vertical $fov$ maps are calculated from the $cc$ maps and also depend on the camera focal length $f$
\begin{equation}
    \label{eq:ff}
    fov_{ch}[i,j] = \arctan\Big(\frac{cc_{ch}[i,j]}{f}\Big),
\end{equation}
where $ch$ can be $x$ or $y$ (see Eq. \ref{eq:ccx} and \ref{eq:ccy}). They give information about the captured context and the focal length. These maps are sensitive to sensor size and focal length. In Figure~\ref{fig:camconvs} we represent $fov$ with a color gradient from {\color{darkpastelgreen}green} to {\color{magenta}pink}; {\color{selectiveyellow}yellow} represents an angle of 0 in the field of view map. Notice in the bottom part of the figure how the $fov$ map values change when changing camera focal length, sensor size or principal point. Changes on the pixel size change the resolution of the map but the field of view and thus the available context in the image stays the same.

\textbf{Normalized Coordinates ($nc$):} We also include a Coord-Conv channel of normalized coordinates  \cite{liu2018intriguing}. The values of Normalized Coordinates vary linearly with the image coordinates between $[-1,1]$. This channel does not depend on the camera sensor. However, it is very useful to describe the spatial extent of the context (in feature space) that is left in each direction (\textit{e.g.}, if the value on the $x$ channel is close to $-1$, it means the feature vector at this position is close to the left border and there is almost no context on the left side).

Notice that $nc$ is not shown in Figure~\ref{fig:camconvs} as it remains constant.

\subsection{Focal Length Normalization}
\label{sec:focalnorm}
An instance of an object imaged by two cameras with different focal length appears with different image sizes although the depth is the same. 
Focal length normalization is an alternative to avoid such inconsistencies.
To this end, we predict depth values normalized to a default focal length $f_n$. Given a metric depth map $d$ we get the normalized depth values as $\frac{f_n}{f}d$ with $f$ as the actual focal length. Note that the normalized depth values depend on the focal length. For the raw inverse depth predictions $\tilde{\xi}$ of our network we denormalize the values as  
\begin{equation}
\label{eq:normalizeddepth}
\xi=\frac{f_{n}}{f}\tilde{\xi},
\end{equation}
where $\xi=\frac{1}{d}$ is the inverse depth map.
\cite{tateno2017cnn} used a similar approach to correct depth values at test time, in this paper we propose for the first time to use it during training. 

This normalization can be used together with our \shortConvs{}. Although \shortConvs{} allows the network to learn this normalization on its own, we found in our experiments that using this normalization accelerates the convergence. It should be remarked, though, that focal length normalization assumes a constant pixel size over the whole image set, and therefore can only be used in such cases. \shortConvs{} are a more general model that overcomes this limitation.
\section{Model and Training}
\subsection{Network Architecture}
\begin{figure}[t!]
\includegraphics[width=\linewidth]{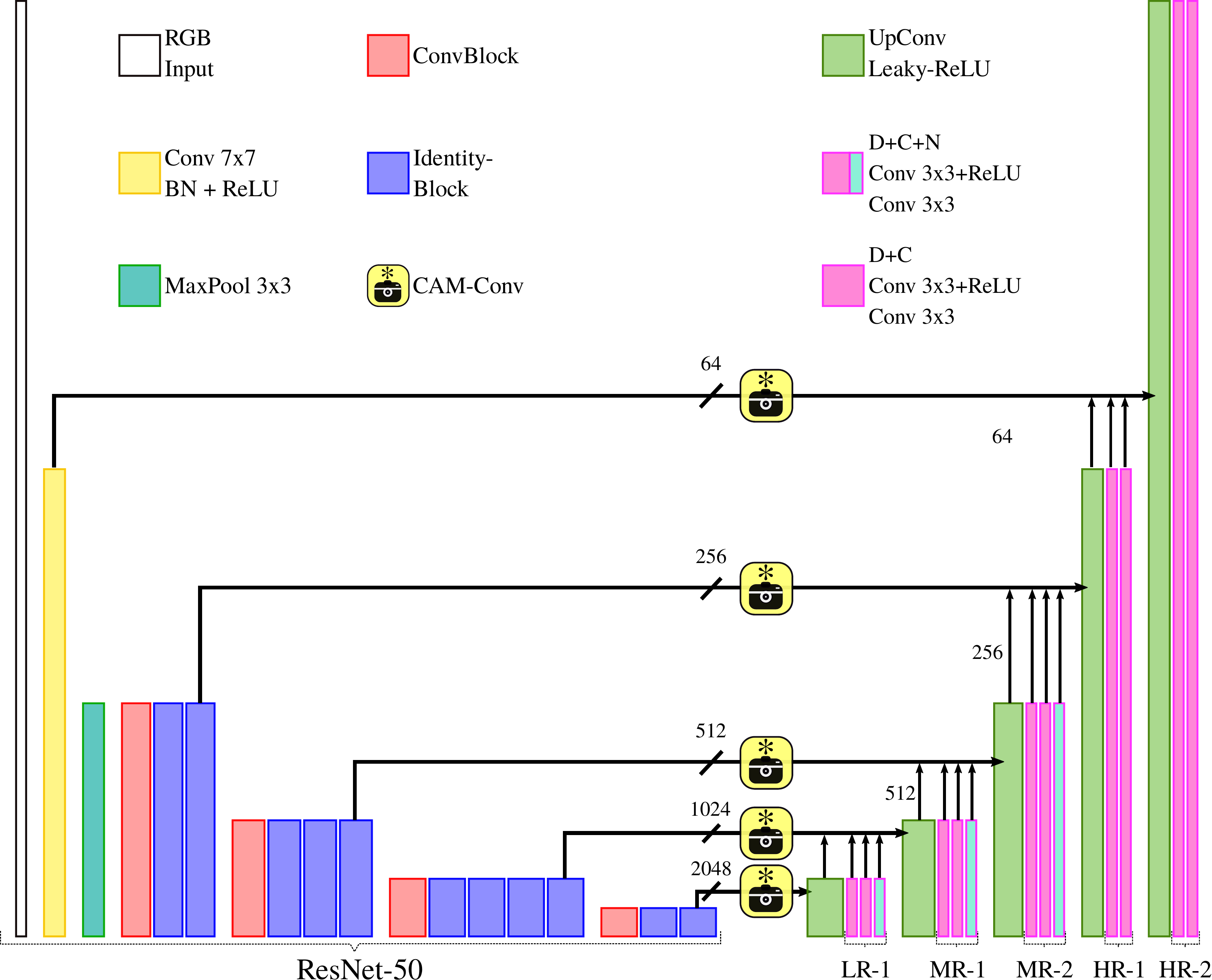}
\caption{\label{fig:ts_network2d}%
Our network architecture, inspired by DispNet \cite{mayer2016large}, to which 
we added \shortConvs{} connecting the encoder and decoder. We predict depth, confidence and normals (D+C+N) in the first three intermediate resolution levels (LR-1,MR-1 and MR-2) and only depth and confidence (D+C) in the last two resolution levels (HR-1 and HR-2).
}
\vspace{-4mm}
\end{figure}
The network we use in this work has an encoder-decoder architecture inspired by DispNet \cite{mayer2016large}. Hence, we add skip-connections from the low-level feature maps of the encoder to the feature maps of the same size in the decoder, and concatenate them \cite{ronneberger2015u}. Withal, we also estimate intermediate pyramid-resolution predictions, which converge faster and ensure that the network's internal features are more aimed for the task. 
As it is common in the literature \cite{laina2016deeper,kuznietsov2017semi}, our network's backbone is ResNet-50, pretrained on the ImageNet Classification Dataset \cite{he2016deep}. As suggested in the literature \cite{godard2018digging} and our experiments, pretraining the encoder on general image recognition tasks, as ImageNet, helps in both accuracy and convergence time reduction. A schematic of our network architecture can be seen in Figure~\ref{fig:ts_network2d}.

The network predictions are composed by:

\textbf{$\xi$: Inverse depth $\xi=\frac{1}{d}$.} We chose inverse depth for its linear relationship with pixel variations. 

\textbf{$\mathit{c}$: Depth confidence.} As \cite{ummenhofer2017demon}, we enforce the network to predict a confidence map for every depth prediction. 

\textbf{$\mathbf{n}$: Surface normals.} The normals are predicted only for small resolutions (all except the last two), as the ground-truth normals are too noisy at full resolution.%

\subsection{Losses}
\label{sec:losses}

In this section we will present all the losses and their combination for the training.

\textbf{Depth Loss:}
We minimize the L1 norm of the predicted inverse depth $\xi$ minus the ground truth inverse depth $\hat{\xi}$, that is
\begin{equation}
\label{eq:depthloss}
\mathcal{L}_{d}=\sum_{i,j}\Big|\xi(i,j)-\hat{\xi}(i,j)\Big|.
\end{equation}

Note that for experiments with focal length normalization we scale depth values accordingly (see section \ref{sec:focalnorm}).

\textbf{Scale-Invariant Gradient Loss:} We use the scale-invariant gradient loss proposed by \cite{ummenhofer2017demon}, in order to favor smooth and edge preserving depth estimations. The loss based on the depths is
\begin{equation}
\label{eq:sinv-grad-loss}
\mathcal{L}_{g}=\sum_{h=\{1,2,4,8,16\}}\sum_{i,j}\Big|\Big|\mathbf{g}_h[\xi](i,j)-\mathbf{g}_h[\hat{\xi}](i,j)\Big|\Big|_2.
\end{equation}

For the gradients, we use the same discrete scale-invariant finite differences operator $\mathbf{g}$ as defined in their work, which is
\begin{equation}
\label{eq:sinv-grad}
\mathbf{g}_h[d](i,j)=\bigg(\text{\footnotesize $\frac{d(i+h,j)-d(i,j)}{|d(i+h,j)+d(i,j)|},\frac{d(i,j+h)-d(i,j)}{|d(i,j+h)+d(i,j)|}$}\bigg)^\top,
\end{equation}
and we apply the scale-invariant loss to cover gradients at 5 different spacings $h$.

\textbf{Confidence Loss:} 
The ground truth for the confidence map must be calculated online as it depends on the prediction. The confidence ground truth is calculated as 
\begin{equation}
\label{eq:confidence-map}
\hat{\mathit{c}}(i,j)=e^{-|\xi(i,j)-\hat{\xi}(i,j)|},
\end{equation}
and its corresponding loss function is defined as
\begin{equation}
\label{eq:confidence-loss}
\mathcal{L}_{c}=\sum_{i,j}\Big|\mathit{c}(i,j)-\hat{\mathit{c}}(i,j)\Big|.
\vspace{-1ex}
\end{equation}

\textbf{Normal Loss:} 
For the normal loss, we use the L2 norm. The ground truth for the normals ($\hat{\mathbf{n}}$) is derived from the ground truth depth image. %
The loss for the normals is as follows:
\begin{equation}
\label{eq:normals-loss}
\mathcal{L}_{n} =\sum_{i,j}\Big|\Big| \mathbf{n}(i,j)-\hat{\mathbf{n}}(i,j) |\Big|\Big|_2.
\end{equation}

\textbf{Total Loss:}  The individual losses are weighted by factors obtained empirically, so the total loss $\mathcal{L}$ is
\begin{equation}
\begin{split}
\label{prediction-loss}
\mathcal{L}_{}=\lambda_1\mathcal{L}_{d}+\lambda_2\mathcal{L}_{g}+\lambda_3\mathcal{L}_{c}+\lambda_4\mathcal{L}_{n},
\end{split}
\end{equation}
where $\lambda_1$, $\lambda_2$, $\lambda_3$ and $\lambda_4$ are $150$, $100$, $50$ and $25$ respectively.
\section{Multi-Camera Experiments and Results}
Most of the single-view depth prediction networks have been trained and tested using the same or very similar camera models. Generalizing to different camera models has several implications that are not straightforward. For this reason, we first present a thorough analysis on the generalization capabilities of current approaches.
To this end we apply na\"ive generalization techniques (focal normalization and image resizing) during training on a network without our special convolutions (as Figure \ref{fig:ts_network2d} but without \shortConvs{}) and examine the limitations.
Finally, we train and evaluate our network with \shortConvs{} (as Figure~\ref{fig:ts_network2d}) and show the improved generalization performance  with respect to different camera parameters.

\subsection{Experimental Setup}
The major part of our experiments are done on the 2D-3D Semantics Dataset \cite{armeni2017joint}, that contains RGB-D equi-rectangular images. 
This dataset allows us to generate images with different camera intrinsics \emph{but} the same content. We have observed that depth estimation networks overfit to the camera parameters and the image content distribution (the latter being different in indoors and outdoors datasets, for example). In this manner we eliminate the content distribution factor and isolate the effect of the camera parameters. 

All the experiments were done using the 3-fold cross-validation suggested in \cite{armeni2017joint}. In this section we present median values for the most relevant experiments. To see the complete results, more details on the dataset and image generation process and additional experiments we refer the reader to the supplementary material.

\begin{table}[t!]
\small
\centering
\begin{tabu}{lcccc}
\cline{1-4}
\textbf{Name} & \bigImR{}& \bigImtR{} & \sqImR{} \\
\cline{1-4}
\rowfont{\scriptsize}{\small\textbf{Sensor}}&   \bigIm{}& \bigImt{} & \sqIm{} \\
\cline{1-4}
\\
\hline
\textbf{Name} &  \smallImR{} &\extraImR{} & \scannetSizeR{} & \kittiSizeR{}\\
\hline
\rowfont{\scriptsize}{\small\textbf{Sensor}}&  \smallIm{} &\extraIm{} & \scannetSize{} & \kittiSize{}\\
\hline
\\
\hline
\textbf{Name} &\smallFocalR{} & \bigFocalR{} & \extraFocalR{} & \normFocalR{}\\ 
\hline
\textbf{Focal} &\smallFocal{} & \bigFocal{} & \extraFocal{} & \normFocal{}\\ 
\hline
\end{tabu}
\caption{\label{tab:taxo} Notation for different sensor sizes and focal lengths.}
\vspace{-4mm}
\end{table}

The notation for sensor sizes and focal lengths used during the evaluation is in Table \ref{tab:taxo}. As an example, if a network has been trained with sensor sizes \bigImt{} and \sqIm, and focal length \smallFocal{}, we will denote this model as \bigImtR{}\sqImR{}\smallFocalR{}. In some experiments we use a random distribution for the focal length. As an example, if the synthesized focal lengths are uniformly distributed between \smallFocal{} and \bigFocal{}, the model will be denoted as $\mathcal{U}$\smallFocalR{}\bigFocalR{}.

We evaluate the performance on both depth and inverse depth. All the error metrics we used in our experiments are standard from the literature. In addition we use relative metrics and the scale-invariant metric presented by \cite{eigen2014depth}, which are widely used in depth estimation.

\subsection{Influence of context}
Modifying the camera parameters affects the field of view, and hence the amount of context the image is capturing. We evaluate the influence of the context in the depth prediction of a standard U-Net encoder-decoder architecture (network in Figure \ref{fig:ts_network2d} without \shortConvs{}) with two different experiments. First, we compare two networks trained with images with sensor size \bigImR{} and two different focal lengths \bigFocalR{} and \extraFocalR{} (Table \ref{tab:context_same_size}). Second, we compare two networks with images with the same focal length but different sensor sizes: \bigImR{}  and \smallImR{} (Table \ref{tab:context_same_focal}). 

As expected, context helps. The performance is better for the smallest focal \extraFocalR{}, which results in a wider FOV and hence more context. Also the performance is better for the bigger sensor size \bigImR{}, which also provides more context. To remove the context dependency in our analysis, for some of the experiments in next subsections we will generate images with uniformly distributed focal lengths.

\begin{table}[t!]
\footnotesize
\centering
\begin{tabular}{@{\extracolsep{4pt}}ccccccc}
\textbf{Test} & \textbf{Train} & abs.rel&rmse&sc.inv&sq.rel\\
\hline
&  & $:1$ & $m$ & $lg(m)$& $:1$\\
\bigImR{}\extraFocalR{}&\bigImR{}\extraFocalR{}&$\mathbf{0.17}$  &$\mathbf{0.378}$ &$\mathbf{0.0347}$ &$\mathbf{0.048}$ \\
\bigImR{}\bigFocalR{} &\bigImR{}\bigFocalR{} &$0.195$ &$0.51$ &$0.0387$ &$0.0606$ \\
\cline{3-6}
\multicolumn{2}{c}{}&\multicolumn{4}{c}{\textit{\stb{}}}
\\
\end{tabular}
\caption{\label{tab:context_same_size}Influence of context, different focal lengths. }
\end{table}
\begin{table}[t!]
\footnotesize
\centering
\begin{tabular}{@{\extracolsep{4pt}}ccccccc}
\textbf{Test} & \textbf{Train}  & abs.rel&rmse&sc.inv&sq.rel\\
\hline
&  & $:1$& $m$ & $lg(m)$&$:1$ \\
\bigImR{}\extraFocalR{}&\bigImR{}\extraFocalR{}&$\mathbf{0.17}$  &$\mathbf{0.378}$ &$\mathbf{0.0347}$ &$\mathbf{0.048}$ \\
\smallImR{}\extraFocalR{} &\smallImR{}\extraFocalR{} &$0.204$ & $0.54$ & $0.0384$ & $0.0637$\\
\cline{3-6}
\multicolumn{2}{c}{}&\multicolumn{4}{c}{\textit{\stb{}}}
\\
\end{tabular}
\caption{\label{tab:context_same_focal}Influence of context, different sensor sizes.}
\vspace{-4mm}
\end{table}
\subsection{Overfitting of standard networks}

In this experiment we evaluate the performance of a standard U-Net architecture for variations of the camera parameters on the training and test sets. We will focus the study on two parameters: (a) focal length and (b) sensor size. First we will fix the sensor size to \bigImR{} and we will test on images with focal lengths \extraFocalR{}, \smallFocalR{} and \bigFocalR{} (first three test sets in Table \ref{tab:overfitting}). Second we will sample random focal lengths from a uniform distribution between \smallFocalR{} and \bigFocalR{} and we will evaluate on images with sensor sizes \bigImR{} and \bigImtR{} (last two test sets in Table \ref{tab:overfitting}). For every test set there are $4$ to $5$ different train sets (referred in the \nth{2} column of the table). For every test set we will refer to the case where the cameras from the training and test set are the same as the {\color{baseline1}same-camera baseline}. 
Training sets where we did not use focal length normalization are denoted with a '*'.
Networks trained on train sets with two sensor sizes have been trained either as Siamese networks with weight sharing or with image resizing to size $s_1$ (denoted with a '\textdagger').

It is important to remark that, for all the experiments, the test and training data was generated from the exact same images and the networks have the same architecture and were trained for the same number of iterations. Any performance variation, then, should be attributed to the variations in the camera intrinsics and the na\"ive solutions we analyze. Notice in Table \ref{tab:overfitting} that, in general, the {\color{baseline1}same-camera baseline} outperforms the rest, demonstrating the overfit to the camera parameters.

\begin{table}[t!]
\centering
\footnotesize
\begin{threeparttable}[b]
\begin{tabular}{@{\extracolsep{4pt}}=c+c+c+c+c}
\textbf{Test set} & \textbf{Train set}&l1.inv&rmse&sc.inv\\
\hline
$pixels$ & $pixels$ & $1/m$ & $m$ & $lg(m)$\\
\rowstyle{\color{baseline1}}
\multirow{ 5}{*}{\scriptsize\bigImR{}\extraFocalR{}} 
  &\multirow{ 1}{*}{\scriptsize\bigImR{}\extraFocalR{}\tnote{*}}
  	   &$\mathbf{0.184}$ &$\mathbf{0.378}$ &$\mathbf{0.0347}$ \\
  &\multirow{ 1}{*}{\scriptsize\bigImR{}\smallFocalR{}}
  	  &$0.193$ &$0.395$ &${0.0354}$\\
  &\multirow{ 1}{*}{\scriptsize\bigImR{}\bigFocalR{}}
  	 &$0.318$ &$0.572$ &$0.0483$ \\
   &\multirow{ 1}{*}{\scriptsize\bigImR{}\smallFocalR{}\bigFocalR{}\tnote{*}}
  	 &$0.659$ &$0.864$ &$0.0614$ \\
   &\multirow{ 1}{*}{\scriptsize\bigImR{}\smallFocalR{}\bigFocalR{}}
  	 &${0.189}$ &${0.387}$ &$0.0361$\\
\hline
\rowstyle{\color{baseline1}}
\multirow{ 4}{*}{\scriptsize\bigImR{}\smallFocalR{}} 
  &\multirow{ 1}{*}{\scriptsize\bigImR{}\smallFocalR{}\tnote{*}}
  	&$\mathbf{0.17}$ &$\mathbf{0.4}$ &$\mathbf{0.0354}$ \\

  &\multirow{ 1}{*}{\scriptsize\bigImR{}\bigFocalR{}}
  	 &$0.272$ &$0.564$ &$0.0459$ \\
   &\multirow{ 1}{*}{\scriptsize\bigImR{}\smallFocalR{}\bigFocalR{}\tnote{*}}
  	&$0.552$ &$0.888$ &$0.0609$\\
   &\multirow{ 1}{*}{\scriptsize\bigImR{}\smallFocalR{}\bigFocalR{}}
  	 &${0.175}$ &${0.404}$ &${0.0364}$ \\

\hline
\rowstyle{\color{baseline1}}
\multirow{ 4}{*}{\scriptsize\bigImR{}\bigFocalR{}} 
  &\multirow{ 1}{*}{\scriptsize\bigImR{}\bigFocalR{}\tnote{*}}
  	 &$0.141$ &$0.51$ &$0.0387$ \\
  &\multirow{ 1}{*}{\scriptsize\bigImR{}\smallFocalR{}}
  	 &$0.133$ &$0.524$ &$0.0411$ \\
   &\multirow{ 1}{*}{\scriptsize\bigImR{}\smallFocalR{}\bigFocalR{}\tnote{*}}
  	 &$0.208$ &$0.813$ &$0.063$ \\
   &\multirow{ 1}{*}{\scriptsize\bigImR{}\smallFocalR{}\bigFocalR{}}
  	 &$\mathbf{0.132}$ &$\mathbf{0.504}$ &$\mathbf{0.038}$ \\

 \hline

\multirow{ 4}{*}{\scriptsize\bigImR{}$\mathcal{U}$\smallFocalR{}\bigFocalR{}} 
\rowstyle{\color{baseline1}}
  &\multirow{ 1}{*}{\scriptsize\bigImR{}$\mathcal{U}$\smallFocalR{}\bigFocalR{}}
  	&$\mathbf{0.15}$ &${\mathbf{0.46}}$ &${\mathbf{0.037}}$\\
   
  &\multirow{ 1}{*}{\scriptsize\bigImtR{}$\mathcal{U}$\smallFocalR{}\bigFocalR{}}
 &$0.175$ &$0.51$ &$0.0422$\\
  &\multirow{ 1}{*}{\scriptsize\centering\bigImR{}\bigImtR{}$\mathcal{U}$\smallFocalR{}\bigFocalR{}}
  	 &$0.153$ &$0.484$ &$0.0401$\\
  &\multirow{ 1}{*}{\scriptsize\centering\bigImR{} \bigImtR{} \sqImR{}$\mathcal{U}$\smallFocalR{}\bigFocalR{}\tnote{\textdagger}} &$0.179$ &$0.742$ &$0.064$\\
 
  \hline

  \multirow{ 4}{*}{\scriptsize\bigImtR{}$\mathcal{U}$\smallFocalR{}\bigFocalR{}} 

  &\multirow{ 1}{*}{\scriptsize\bigImR{}$\mathcal{U}$\smallFocalR{}\bigFocalR{}}
  	 &$0.151$ &$0.44$ &$0.038$ \\
    \rowstyle{\color{baseline1}}
  &\multirow{ 1}{*}{\scriptsize\bigImtR{}$\mathcal{U}$\smallFocalR{}\bigFocalR{}}
 &${\mathbf{0.133}}$ &$\mathbf{0.412}$ &$\mathbf{0.0323}$\\
  &\multirow{ 1}{*}{\scriptsize\centering\bigImR{}\bigImtR{}$\mathcal{U}$\smallFocalR{}\bigFocalR{}}
  	 &$0.139$ &$0.436$ &$0.0352$\\
  &\multirow{ 1}{*}{\scriptsize\centering\bigImR{} \bigImtR{} \sqImR{}$\mathcal{U}$\smallFocalR{}\bigFocalR{}\tnote{\textdagger}}	 &$0.16$ &$0.622$ &$0.0514$\\
\cline{3-5}
\multicolumn{2}{c}{}&\multicolumn{3}{c}{\stb{}}\\
\end{tabular}
\begin{tablenotes}
  \item[*] trained without focal length normalization.
  \item[\textdagger] images resized to \bigImR{} during training.
\end{tablenotes}
\end{threeparttable}
\\
\caption{\label{tab:overfitting}Overfit to camera parameters of standard encoder-decoder architectures. Networks trained from images with variations in their intrinsics perform worse than the {\color{baseline1}same-camera baseline}.}
\vspace{-4mm}
\end{table}

The conclusions of these experiments are as follows.

\noindent\textbf{(a) Single-focal training overfits. }The performance of a depth network degrades when trained on images from a particular camera and tested on images from different cameras. See, for example, the drop in performance between the \nth{1} row (test:~$s_1f_{64}$, train: {\color{baseline1}${s_1f_{64}}^*$}) and the \nth{2} (test:~$s_1f_{64}$, train: ${s_1f_{72}}$) and \nth{3} (test:~$s_1f_{64}$, train: ${s_1f_{128}}$) rows in all metrics.%

\noindent\textbf{Multi-focal training with normalization helps. }The results improve when the training set contains images with different focal lengths and is done with focal normalization. See, for example, that the results on test set \bigImR{}\extraFocalR{} with training set \bigImR{}\smallFocalR{}\bigFocalR{} is close to the {\color{baseline1}same-camera baseline}. Notice, however, that the multi-focal train set does not reach the performance of the {\color{baseline1}same-camera baseline}. In section \ref{sec:expcamconvs} we will show how \shortConvs{} are able to outperform the {\color{baseline1}same-camera baseline} even when the training data does not contain the test focal length.

The performance degrades without focal normalization. Compare, for example, the error metrics of the train sets \bigImR{}\smallFocalR{}\bigFocalR{}$^*$ and \bigImR{}\smallFocalR{}\bigFocalR{}. Networks trained on \smallFocalR{}\bigFocalR{}$^*$, in fact, did not converge easily. 

\noindent\textbf{Limitations of focal normalization. }Two things should be noticed regarding focal normalization: First, it does not model the changes on the sensor size and the resolution, and we will see now how changes on them degrade the performance. 
And second, Equation \ref{eq:normalizeddepth} only holds if the pixel size is the same for every camera in the training and test sets, which in general is not the case.

\noindent\textbf{(b) Single-sensor size training overfits. } Networks trained on a sensor size and tested on other sensor sizes do not perform as well as the {\color{baseline1}same-camera baseline}. 
This can be seen in Table~\ref{tab:overfitting} in the last two test sets $s_1\mathcal{U}f_{72}f_{128}$ and $s_2\mathcal{U}f_{72}f_{128}$.
Single-view depth estimation is a context-dependent task, and the network overfits to the amount of context in the training sensor size. 

\noindent\textbf{Multi-sensor size training with weight sharing does not generalize. }Training with multiple sensor sizes works better than training with the wrong sensor size but cannot reach the same performance as {\color{baseline1}same-camera baselines}.
Further, training a stack of weight sharing networks also does not scale to large numbers of different sensor sizes.

\noindent\textbf{Resizing does not work. } As a na\"ive approach, which scales to multiple sensor sizes, we use resizing (denoted with '\textdagger' in Table~\ref{tab:overfitting}), which converts all the images to size (\bigImR{}) during training. Notice that resizing changes the aspect ratio. It also implies the recalculation of a new average focal length $f_r=f\frac{r_x+r_y}{2}$ for normalization. The performance degradation introduced by resizing is noticeable. Resizing creates inconsistent data in train and testing, which leads to learning and convergence difficulties.

\begin{table}
\footnotesize
\centering
\begin{threeparttable}[b]
\begin{tabular}{=c+c+c+c+c+c}
\textbf{Test} & \textbf{Train}& abs.rel&rmse.inv&sc.inv&sq.rel \\
\hline
 &&$\%$&$1/km$&$lg(m)100$&$\%$\\

\multirow{ 3}{*}{\kittiSizeR{}} 
\rowstyle{\color{baseline1}}
  &\multirow{ 1}{*}{\kittiSizeR{}}
  	& $9.16$ &$\mathbf{10.54}$ & $\mathbf{13.3}$ & $\mathbf{2.33}$  \\
  &\multirow{ 1}{*}{\centering \scannetSizeR{}\kittiSizeR{}}
  	& $24.58$ &$36.82$ & $26.51$ & $9.28$  \\
  &\multirow{ 1}{*}{\centering \scannetSizeR{}\kittiSizeR{}\tnote{\textdagger}}
  	& $\mathbf{9.08}$ &$10.55$ & $13.98$ & $2.56$ \\
    \\
 &&abs.rel&l1.inv&rmse.inv&sq.rel \\
\hline
 &&$:1$&$1/m$&$1/m$&$:1$\\
\multirow{ 3}{*}{\scannetSizeR{}} 
\rowstyle{\color{baseline1}}
  &\multirow{ 1}{*}{\scannetSizeR{}}
  	& $\mathbf{0.12	}$ &$\mathbf{0.09}$ & $\mathbf{0.12}$ & $\mathbf{0.03}$\\
  &\multirow{ 1}{*}{\centering \scannetSizeR{}\kittiSizeR{}}
  	& $0.26$ &$0.13$ & $0.16$ & $0.18$\\
  &\multirow{ 1}{*}{\centering \scannetSizeR{}\kittiSizeR{}\tnote{\textdagger}}
  	& $\mathbf{0.12}$ &$\mathbf{0.09}$ & $\mathbf{0.12}$ & $\mathbf{0.03}$\\
\cline{3-6} 
\multicolumn{2}{c}{}&\multicolumn{4}{c}{\stb{}}
\\
\end{tabular}
\begin{tablenotes}
  \item[\textdagger] sensor size has been resized to the first one in the list.
\end{tablenotes}
\end{threeparttable}
\\
\caption{\label{tab:diffsizekittiscan} Na\"ive train and test on KITTI \cite{Uhrig2017THREEDV} and ScanNet \cite{dai2017scannet}. See that training FCN in multiple image sizes (\kittiSizeR{}\scannetSizeR{}) does not generalize. Resizing works, but only in this particular case, because of the small overlap of visual features.}
\vspace{-5mm}
\end{table}

\noindent\textbf{Resizing helps only in a particular case} (non-overlapping distributions of visual features). Table \ref{tab:diffsizekittiscan} shows an experiment, similar to the previous one, on two public datasets: KITTI \cite{Uhrig2017THREEDV}, with sensor size \kittiSizeR{}, and ScanNet \cite{dai2017scannet}, with sensor size \scannetSizeR{}. In this case, training with both sensor sizes (by weight sharing) decreased the performance. However, resizing reduced the error to the level of the {\color{baseline1}same-camera baselines}. The reason for this is the completely different distribution of the two datasets, with null intersection of visual features (\textit{e.g.} there are no chairs on KITTI and no cars on ScanNet). This is, however, a very particular case, resizing degrades significantly the accuracy in general.

\subsection{Robust Generalization with \shortConvs{}}
\label{sec:expcamconvs}

In this experiment we show that \shortConvs{} generalize to different camera models. In order to evaluate the influence of \shortConvs{} we trained our model with two different sensor sizes (\bigImR{} and \bigImtR{}) and weight sharing. Focal length during training is sampled randomly from a uniform distribution $\mathcal{U}$\smallFocalR{}\bigFocalR{}. We evaluated the trained model in four different test sets, see Table \ref{tab:camconvs}. The first two include the camera model the network was trained with, the third has a sensor size unseen during training, and the last ($s_5f_{64}$) was generated from a camera completely different from the training ones with bigger sensor size and smaller focal length. This case augments considerably the context--\emph{e.g} field of view--which proved to be the hardest case in previous experiments (see network trained with \bigImR{}\bigFocalR{} in Table \ref{tab:overfitting}).

\begin{table}
\scriptsize
\centering
\begin{threeparttable}[b]
\begin{tabular}{=c@{\hskip 1px}+c+c+c+c+c+c+c+c+c}
\textbf{Test} & \textbf{Train}& abs.rel&l1.inv&rmse&sc.inv  \\
\hline
 &  & $:1$ & $1/m$ & $m$ & $lg(m)$ \\

\multirow{ 2}{*}{\bigImR{}$\mathcal{U}$\smallFocalR{}\bigFocalR{}} 
\rowstyle{\color{baseline1}}
  &\multirow{ 1}{*}{\bigImR{}$\mathcal{U}$\smallFocalR{}\bigFocalR{}}
  	&$0.189$ &$0.15$ &$0.46$ &$0.037$\\
  &\multirow{ 1}{*}{\withBoxNet{}\tnote{\ddag}}  
    &$\mathbf{0.175}$ &$\mathbf{0.144}$ &$\mathbf{0.433}$ &$\mathbf{0.0312}$ \\ 
    
  \hline

 \multirow{ 2}{*}{\bigImtR{}$\mathcal{U}$\smallFocalR{}\bigFocalR{}}  
    \rowstyle{\color{baseline1}}
  &\multirow{ 1}{*}{\bigImtR{}$\mathcal{U}$\smallFocalR{}\bigFocalR{}}
&$0.166$ &$0.133$ &$0.412$ &$0.0323$\\
&\multirow{ 1}{*}{\withBoxNet{}\tnote{\ddag}} 
    &$\mathbf{0.158}$ &$\mathbf{0.131}$ &$\mathbf{0.39}$ &$\mathbf{0.0265}$\\
    
  \hline
  
  \rowstyle{\color{baseline1}}
  \multirow{ 5}{*}{\sqImR{}$\mathcal{U}$\smallFocalR{}\bigFocalR{}} 
  &\multirow{ 1}{*}{\sqImR{}$\mathcal{U}$\smallFocalR{}\bigFocalR{}}
  	&$0.174$ &$0.14$ &$0.425$ &$0.0336$\\
  	  &\multirow{ 1}{*}{\bigImR{}$\mathcal{U}$\smallFocalR{}\bigFocalR{}}
  	&$0.184$ &$0.143$ &$0.44$ &$0.0357$\\
   
  &\multirow{ 1}{*}{\bigImtR{}$\mathcal{U}$\smallFocalR{}\bigFocalR{}}
&$0.177$ &$0.145$ &$0.435$ &$0.0356$\\
    &\multirow{ 1}{*}{\centering\bigImR{}\bigImtR{}$\mathcal{U}$\smallFocalR{}\bigFocalR{}}
  	&$0.178$ &$0.143$ &$0.451$ &$0.0365$\\
    &\multirow{ 1}{*}{\withBoxNet{}\tnote{\ddag}} 
    &$\mathbf{0.164}$ &$\mathbf{0.134}$ &$\mathbf{0.402}$ &$\mathbf{0.0283}$\\
   \hline
   
   \rowstyle{\color{baseline1}}
  \multirow{ 4}{*}{\extraImR{}\extraFocalR{}} &\multirow{ 1}{*}{\centering\extraImR{}\extraFocalR{}}
  	&$\mathbf{0.163}$ &$\mathbf{0.227}$ &$0.309$ &$\mathbf{0.0356}$\\
  	
  &\multirow{ 1}{*}{\centering\bigImR{}\extraFocalR{}} 
  	&$0.245$ &$0.292$ &$0.337$ &$0.0598$\\	
  	
  &\multirow{ 1}{*}{\centering\bigImR{}\bigImtR{}$\mathcal{U}$\smallFocalR{}\bigFocalR{}}
    &$0.369$ &$0.369$ &$0.44$ &$0.0427$	\\

  &\multirow{ 1}{*}{\withBoxNet{}\tnote{\ddag}}  
    &$0.177$ &$0.236$ &$\mathbf{0.289}$ &$0.0362$ \\

  	\cline{3-7}
  \multicolumn{2}{c}{}&\multicolumn{4}{c}{\stb{}}
  
\end{tabular}
\begin{tablenotes}
    \item[\ddag] Trained with weight sharing in sensor sizes  \bigImR{}, \bigImtR{} and $\mathcal{U}$\smallFocalR{}\bigFocalR{}.
\end{tablenotes}
\end{threeparttable}
\\
\caption{\label{tab:camconvs} Camera parameter generalization with \shortConvs{}. Results on training and testing on different cameras. $1^{st}$ column: camera parameters for test set. $2^{nd}$ column: camera parameters seen during training. This is a continuation of Table \ref{tab:overfitting}. Notice how the network with \shortConvs{} is the only model that generalize getting better performance than the {\color{baseline1}same-camera baseline} on most test sets.}
\vspace{-3mm}
\end{table}

\noindent\textbf{\shortConvs{} generalize over camera intrinsics, outperforming the {\color{baseline1}same-camera baseline}.} Results on the test sets \bigImR{}$\mathcal{U}$\smallFocalR{}\bigFocalR{} and \bigImtR{}$\mathcal{U}$\smallFocalR{}\bigFocalR{} in Table \ref{tab:camconvs} show that the network with \shortConvs{} trained on images of two sizes clearly outperforms the baselines, which was trained on the exact test size. The addition of \shortConvs{} allowed the network to learn the dependence of the image features from the calibration parameters.

\noindent\textbf{\shortConvs{} generalize to sensor sizes unseen during training.} 
Remarkably, the network with \shortConvs{} also outperforms the {\color{baseline1}same-camera baseline} on the test set with sensor size \sqImR{} (third test set in Table \ref{tab:camconvs}), which is not included in the training data.
Further, it generalizes better than a network trained on the exact same conditions but without \shortConvs{} (see \bigImR{}\bigImtR{}$\mathcal{U}$\smallFocalR{}\bigFocalR{} in the table).

\begin{figure}[t!]
\centering
\includegraphics[width=1.\linewidth]{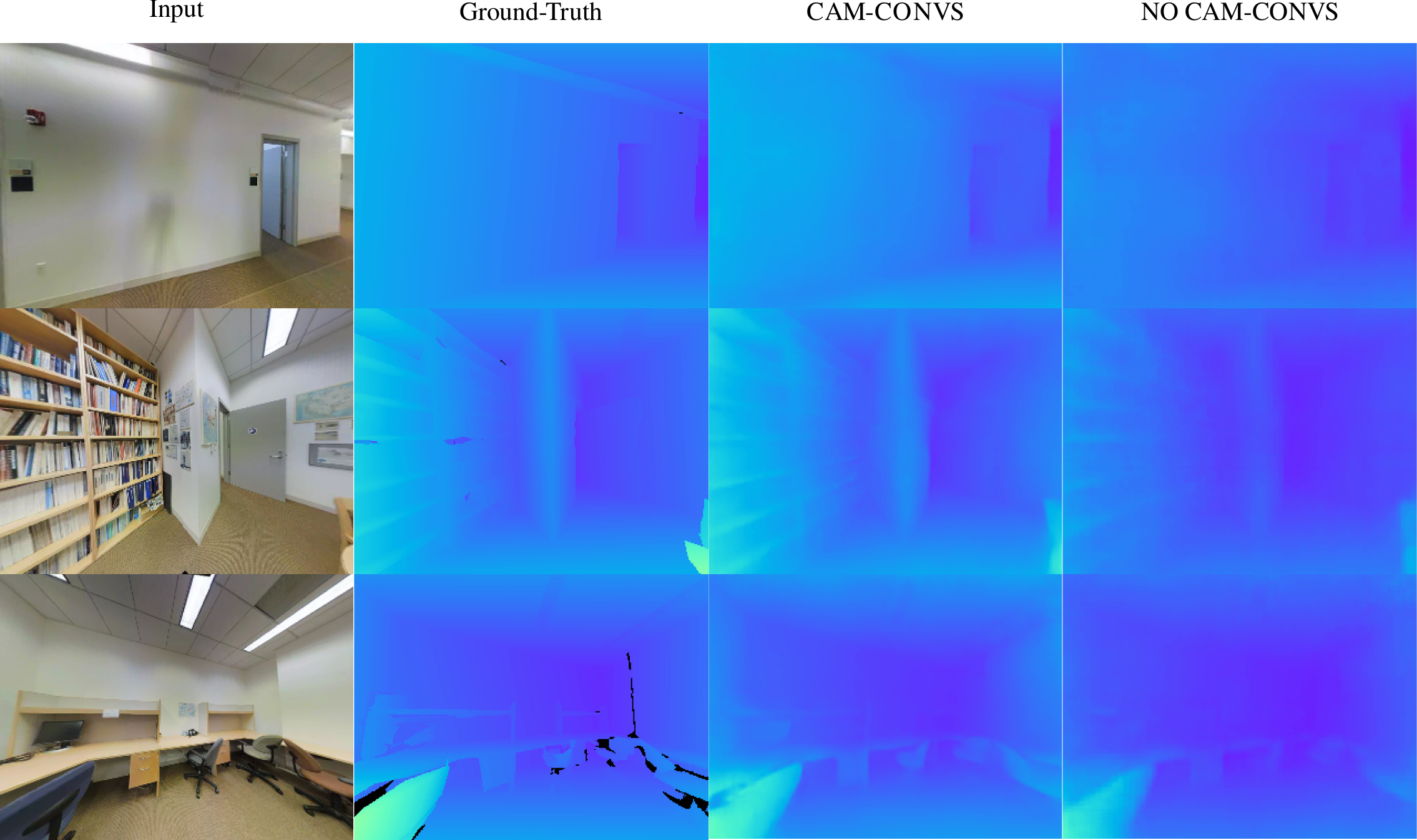}
\caption{\label{fig:withandwithout}  Qualitative results for the test set $s_5f_{64}$. \textbf{\nth{1} column}: RGB input.
\textbf{\nth{2} column:} Ground truth depth. 
\textbf{\nth{3} column:} Prediction with our network using \shortConvs{} trained on $s_1s_2\mathcal{U}f_{72}f_{128}$.
 \textbf{\nth{4} column:} Prediction of a network trained without \shortConvs{}.
Notice that the test camera parameters are significantly different from the training set and images have a much wider field of view. Despite the large difference in the camera parameters the network with \shortConvs{} produces  sharp depth maps on which room corners are clearly visible.
}
\vspace{-3mm}
\end{figure}

\noindent\textbf{\shortConvs{} generalize to cameras unseen during training.} With the last test set (\extraImR{}\extraFocalR{}) in  Table~\ref{tab:camconvs} we evaluate our network on an extreme case of camera parameters with a very wide field of view and very different sensor size from the training ones. Table~\ref{tab:camconvs} shows that \shortConvs{} improve considerably the generalization to new unseen cameras over the na\"ive approaches. 
Figure \ref{fig:withandwithout} shows a qualitative comparison between our network with \shortConvs{} and the network without \shortConvs{} (\bigImR{}\bigImtR{}$\mathcal{U}$\smallFocalR{}\bigFocalR{}) in the test set \extraImR{}\extraFocalR{}.

\subsection{Experiments on Multiple Datasets}
In our last experiment we demonstrate how \shortConvs{} can generalize across datasests by training on four datasets with different cameras  (KITTI \cite{Uhrig2017THREEDV}, ScanNet\cite{dai2017scannet}, MegaDepth \cite{MegaDepthLi18} and Sun3D\cite{xiao2013sun3d} and testing on a different one (NYUv2 \cite{silberman2012indoor}).

\textbf{Training: } 
We trained our network for three different sensor sizes (\oneBsize{}, \twoBsize{} and \threeBsize{}) using weight sharing.
We augmented the training data by scaling the images and shifting the principal point to increase the variation of the camera parameters and then crop to image to one of the target sensor sizes.
We did not use focal length normalization in this experiment, as we cannot ensure constant pixel size across datasets. As MegaDepth has only up-to-scale ground truth, we applied only scale-invariant losses and added the scale-invariant cost function of \cite{eigen2014depth}. %
The same network \emph{without} \shortConvs{}, and hence with no camera information, did not converge during training. The lack of calibration information creates inconsistencies ({\em{}e.g.} same-size objects may have different depths due to different focal lengths).

\begin{figure}[t!]
\centering
\includegraphics[width=1.\linewidth]{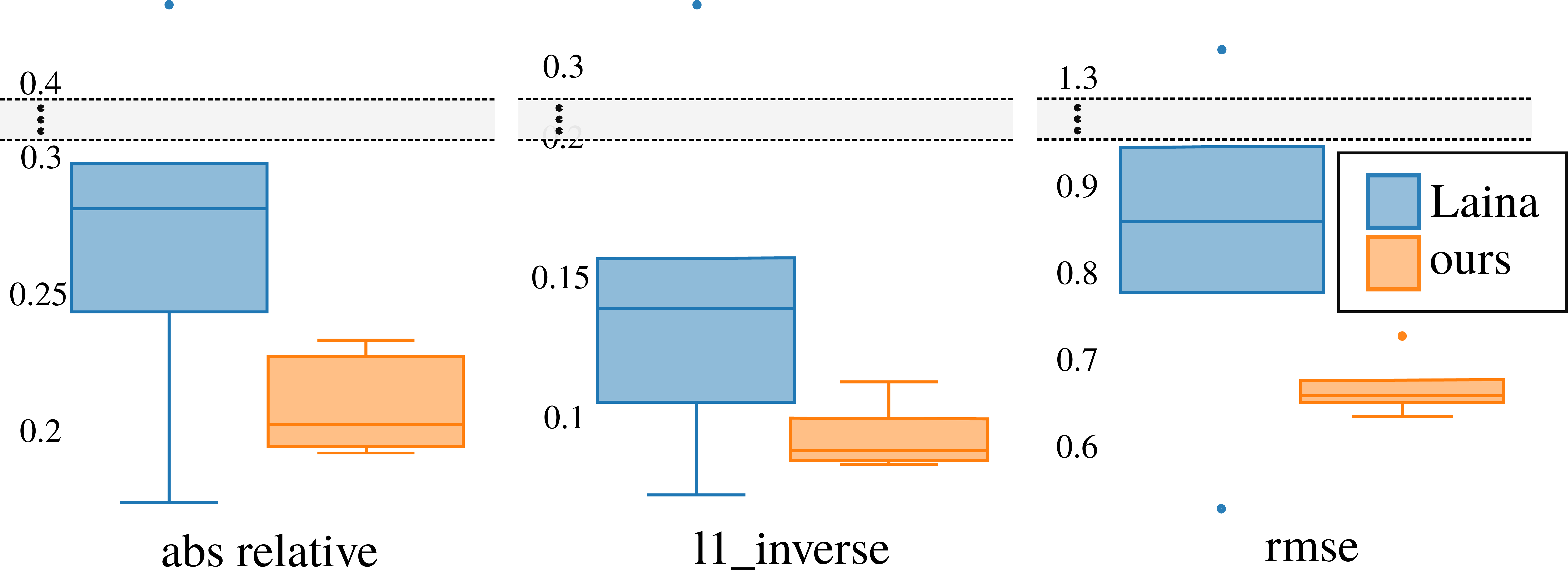}
\caption{\label{fig:nyubox} Error distribution on the test set of NYUv2 with 6 different camera parameters. In orange, our network with \shortConvs{}, trained on several datasets not including NYUv2. In blue, Laina \cite{laina2016deeper}, trained on NYUv2.}
\vspace{-2pt}
\end{figure}

\begin{figure}[t!]
\centering
\includegraphics[width=1.\linewidth]{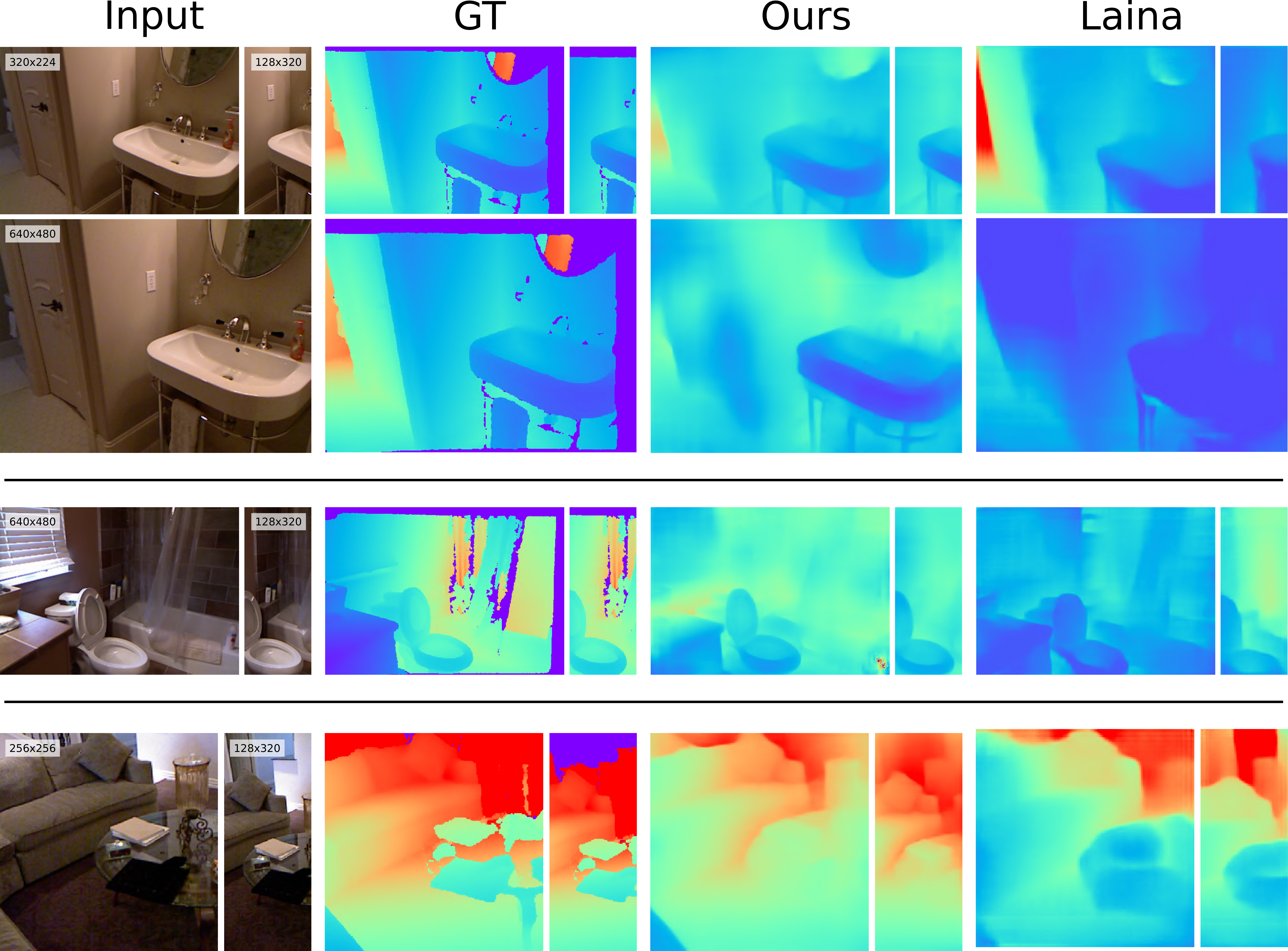}
\caption{\label{fig:nyuqua} Qualitative results, NYUv2 test set with intrinsics variations. 
\textbf{\nth{1} column}: Input RGB images. Each row shows the original one and scaled and cropped versions %
\textbf{\nth{2} column}: Depth groundtruth.
\textbf{\nth{3} column}: Prediction from our network with \shortConvs{}, trained on several datasets \emph{NOT} including NYUv2.
Our network produces consistent depth close to the ground truth for all images.
\textbf{\nth{4} column}: Laina \cite{laina2016deeper}, trained exclusively on NYUv2. Its errors are low on the training resolution but does not generalize to new intrinsics.}
\vspace{-4mm}
\end{figure}

\begin{figure}[t!]
\centering
\includegraphics[width=1.\linewidth]{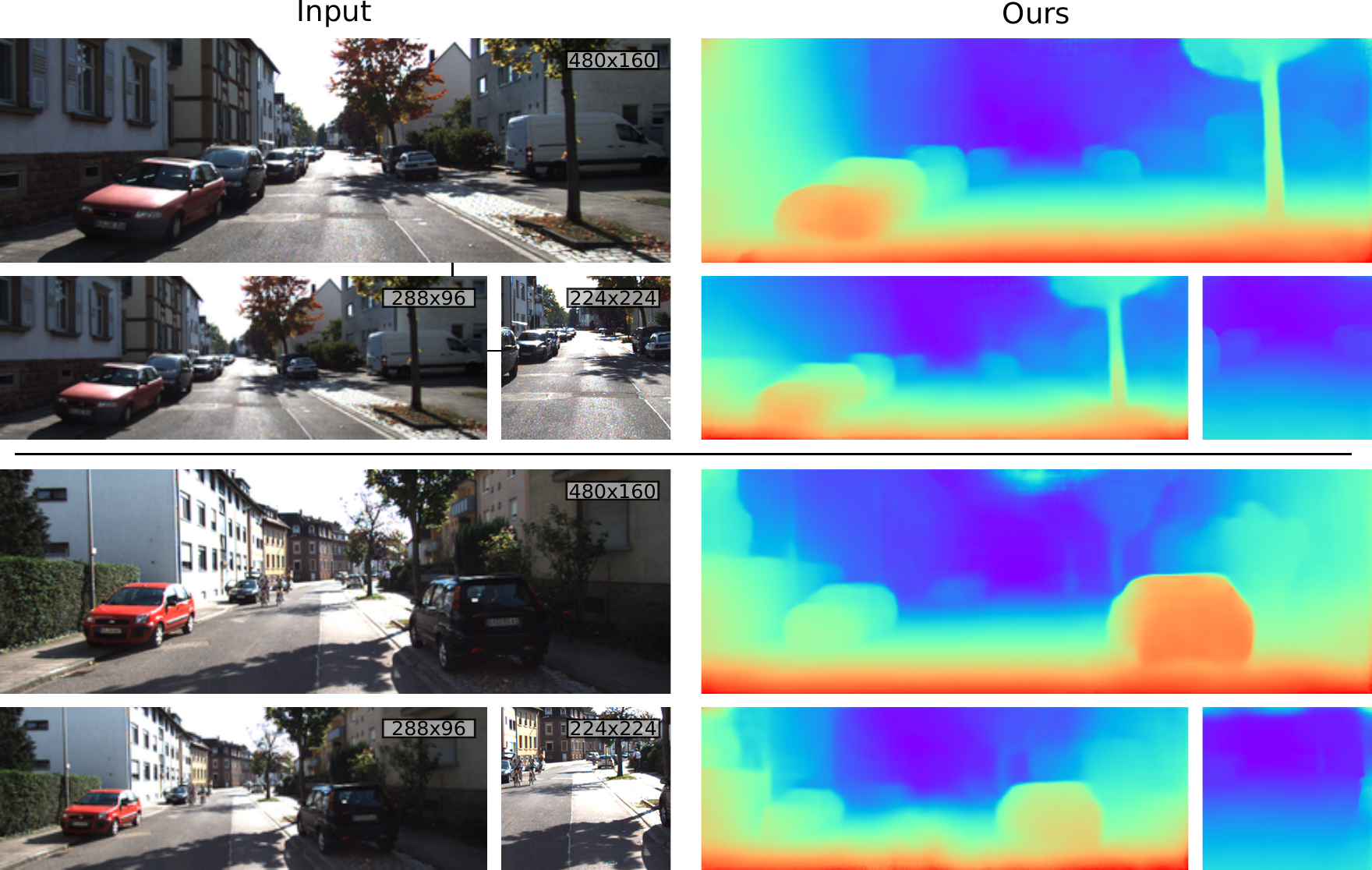}
\caption{\label{fig:kittiqua} Qualitative results on the KITTI validation set. 
\textbf{\nth{1} column:} Input RGB images. Each row shows the original one and scaled and cropped versions. %
\textbf{\nth{2} column:} Prediction from our network.}
\vspace{-3pt}
\end{figure}

\begin{figure}[t!]
\centering
\includegraphics[width=1.\linewidth]{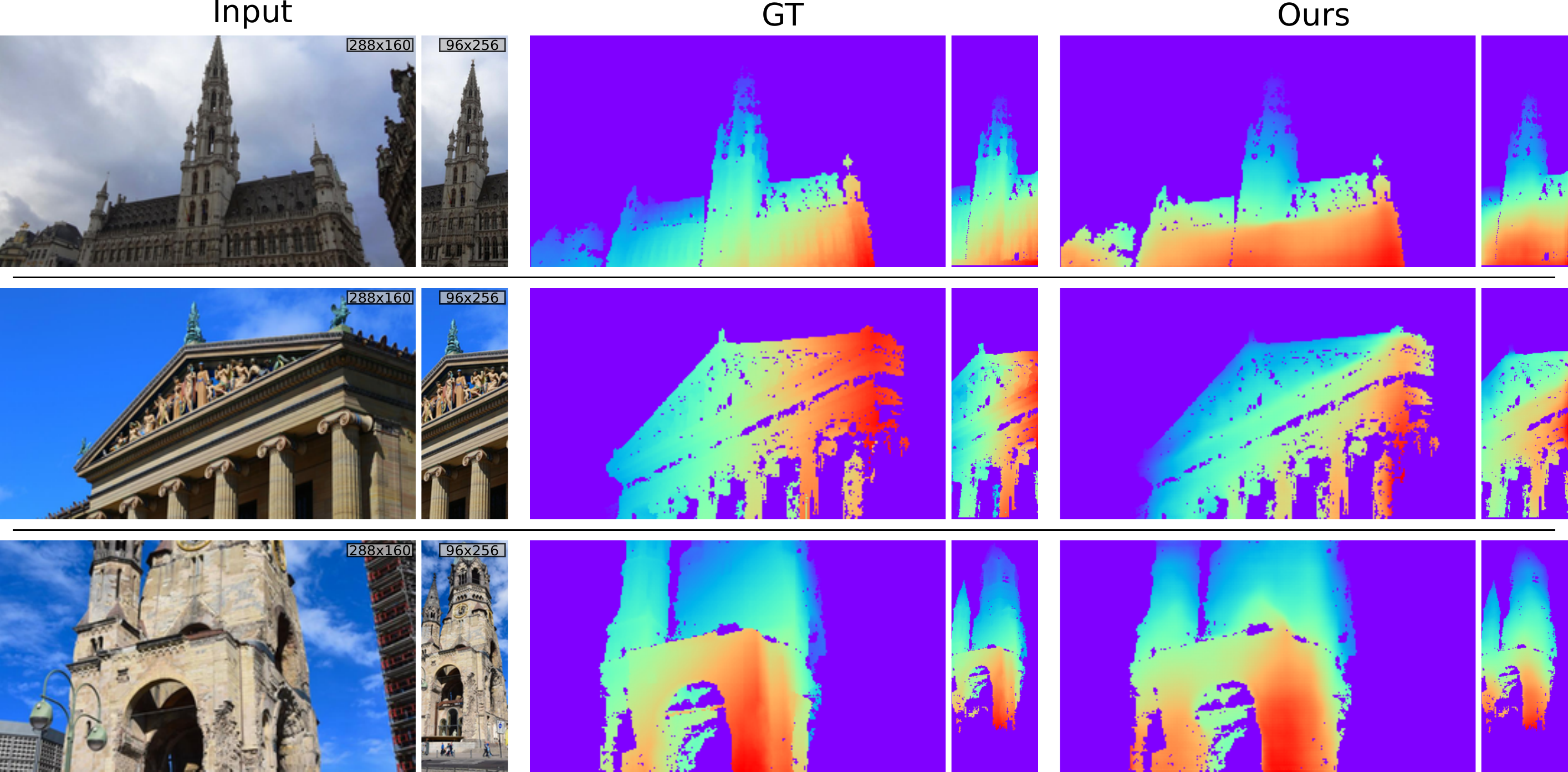}
\caption{\label{fig:megaqua} Qualitative results on MegaDepth test set. 
\textbf{\nth{1} column:} Input RGB images. Each row shows the original one and scaled and cropped versions. %
\textbf{\nth{2} column:} Depth groundtruth.
\textbf{\nth{3} column:} Prediction from our network. The predictions are masked as the groundtruth to facilitate visualization.}
\vspace{-3mm}
\end{figure}

\textbf{Testing: }We evaluated our network on the official test set of NYUv2 and compared against the state of art \cite{laina2016deeper} (similar network without \shortConvs{}) . Note that the network of \cite{laina2016deeper} was trained exclusively on NYUv2, while our network was trained on a set of datasets excluding NYUv2 with different cameras and data distributions (some of the datasets are outdoors, see Figure~\ref{fig:kittiqua} and Figure~\ref{fig:megaqua}). This is important since our model cannot benefit from the dataset bias \cite{torralba2011unbiased}. We predicted depths for images from 6 different cameras: the original camera of the NYUv2 dataset and 5 simulated ones by cropping (to shift principal point and reduce sensor size) and resizing (to change focal length).

Figure \ref{fig:nyubox} shows the distribution of the mean error of the usual metrics obtained for the 6 different cameras. Since  \cite{laina2016deeper} was trained on the NYUv2 dataset, it works slightly better when it predicts the images from the camera it was trained on (the point with the smallest error). However, performance degrades when the camera changes and \shortConvs{} have always smaller error and variance. 
Figure~\ref{fig:nyuqua} illustrate how \shortConvs{} depth predictions are stable for different cameras, while  predictions of \cite{laina2016deeper} vary significantly. %
Recall that \shortConvs{} were not trained on NYUv2, which indicates that they are able to generalize over different camera models and outperform \cite{laina2016deeper} although they trained on the same dataset. 

Figures~\ref{fig:nyuqua}, \ref{fig:kittiqua} and \ref{fig:megaqua} show depth predictions for images (and cropped/resized versions) from the NYUv2, KITTI and MegaDepth test sets. Again, note the excellent performance across datasets with different data distributions and camera intrinsics. All predictions were done with the exact same network without further fine-tuning to a particular dataset or camera parameters. 
\vspace{-1mm}
\section{Conclusions}

This paper introduces \shortConvs{}, a novel type of convolution that allows depth prediction networks to be camera-independent. Experimental results show that current networks overfit to the training camera model resulting on: 1) a lack of generalization to images from other cameras and 2) degraded performance when trained with images from different cameras. 
\shortConvs{} learn how to use the camera intrinsics jointly with the image features to predict depth; solving both limitations. They maintain prediction accuracy for new cameras and better exploit training data from different cameras. 
The latter is an interesting direction to scale up systems that depend on camera parameters.

{
\vspace{1ex}
\noindent\textbf{Acknowledgement: }This project was in part funded by the Spanish government (DPI2015-67275), the EU Horizon 2020 project Trimbot2020, the Arag{\'o}n government (DGA-T45\_17R/FSE) and Fundaci{\'o}n CAI-Ibercaja. We also thank Facebook for their P100 server donation and gift funding; and Nvidia for their Titan X and Xp donation.
}

{\small
\bibliographystyle{ieee}
\bibliography{mabib}

\begin{thebibliography}{10}\itemsep=-1pt

\bibitem{armeni2017joint}
I.~Armeni, S.~Sax, A.~R. Zamir, and S.~Savarese.
\newblock {Joint 2D-3D-semantic data for indoor scene understanding}.
\newblock {\em arXiv preprint arXiv:1702.01105}, 2017.

\bibitem{bloesch2018codeslam}
M.~Bloesch, J.~Czarnowski, R.~Clark, S.~Leutenegger, and A.~J. Davison.
\newblock {CodeSLAM-Learning a Compact, Optimisable Representation for Dense
  Visual SLAM}.
\newblock In {\em Proceedings of the IEEE Conference on Computer Vision and
  Pattern Recognition}, 2018.

\bibitem{chakrabarti2016depth}
A.~Chakrabarti, J.~Shao, and G.~Shakhnarovich.
\newblock Depth from a {Single} {Image} by {Harmonizing} {Overcomplete} {Local}
  {Network} {Predictions}.
\newblock In {\em Proceedings of the 30th {International} {Conference} on
  {Neural} {Information} {Processing} {Systems}}, {NIPS}'16, pages 2666--2674,
  USA, 2016. Curran Associates Inc.

\bibitem{chen2016single-image}
W.~Chen, Z.~Fu, D.~Yang, and J.~Deng.
\newblock Single-{Image} {Depth} {Perception} in the {Wild}.
\newblock In D.~D. Lee, M.~Sugiyama, U.~V. Luxburg, I.~Guyon, and R.~Garnett,
  editors, {\em Advances in {Neural} {Information} {Processing} {Systems} 29},
  pages 730--738. Curran Associates, Inc., 2016.

\bibitem{dai2017scannet}
A.~Dai, A.~X. Chang, M.~Savva, M.~Halber, T.~Funkhouser, and M.~Nie{\ss}ner.
\newblock {ScanNet: Richly-annotated 3D Reconstructions of Indoor Scenes}.
\newblock In {\em Proc. Computer Vision and Pattern Recognition (CVPR), IEEE},
  2017.

\bibitem{eigen2015predicting}
D.~Eigen and R.~Fergus.
\newblock Predicting depth, surface normals and semantic labels with a common
  multi-scale convolutional architecture.
\newblock In {\em Proceedings of the IEEE International Conference on Computer
  Vision}, pages 2650--2658, 2015.

\bibitem{eigen2014depth}
D.~Eigen, C.~Puhrsch, and R.~Fergus.
\newblock Depth map prediction from a single image using a multi-scale deep
  network.
\newblock In {\em Advances in neural information processing systems}, pages
  2366--2374, 2014.

\bibitem{facil2017single}
J.~M. F{\'a}cil, A.~Concha, L.~Montesano, and J.~Civera.
\newblock {Single-View and Multi-View Depth Fusion}.
\newblock {\em IEEE Robotics and Automation Letters}, 2(4):1994--2001, 2017.

\bibitem{fan2017point}
H.~Fan, H.~Su, and L.~J. Guibas.
\newblock A point set generation network for 3d object reconstruction from a
  single image.
\newblock In {\em CVPR}, pages 2463--2471, 2017.

\bibitem{fu2018deep}
H.~Fu, M.~Gong, C.~Wang, K.~Batmanghelich, and D.~Tao.
\newblock Deep ordinal regression network for monocular depth estimation.
\newblock In {\em Proceedings of the IEEE Conference on Computer Vision and
  Pattern Recognition}, pages 2002--2011, 2018.

\bibitem{geiger2012kitti}
A.~Geiger, P.~Lenz, and R.~Urtasun.
\newblock Are we ready for autonomous driving? the kitti vision benchmark
  suite.
\newblock In {\em Computer {Vision} and {Pattern} {Recognition} ({CVPR}), 2012
  {IEEE} {Conference} on}, pages 3354--3361. IEEE, 2012.

\bibitem{godard2018digging}
C.~Godard, O.~Mac~Aodha, and G.~Brostow.
\newblock Digging into self-supervised monocular depth estimation.
\newblock {\em arXiv preprint arXiv:1806.01260}, 2018.

\bibitem{he2016deep}
K.~He, X.~Zhang, S.~Ren, and J.~Sun.
\newblock Deep residual learning for image recognition.
\newblock In {\em Proceedings of the IEEE conference on computer vision and
  pattern recognition}, pages 770--778, 2016.

\bibitem{he2018learning}
L.~He, G.~Wang, and Z.~Hu.
\newblock Learning depth from single images with deep neural network embedding
  focal length.
\newblock {\em IEEE Transactions on Image Processing}, 27(9):4676--4689, 2018.

\bibitem{hoiem2005automatic}
D.~Hoiem, A.~A. Efros, and M.~Hebert.
\newblock Automatic photo pop-up.
\newblock {\em ACM transactions on graphics (TOG)}, 24(3):577--584, 2005.

\bibitem{huang2018deepmvs}
P.-H. Huang, K.~Matzen, J.~Kopf, N.~Ahuja, and J.-B. Huang.
\newblock Deepmvs: Learning multi-view stereopsis.
\newblock In {\em Proceedings of the IEEE Conference on Computer Vision and
  Pattern Recognition}, pages 2821--2830, 2018.

\bibitem{kehl2017ssd}
W.~Kehl, F.~Manhardt, F.~Tombari, S.~Ilic, and N.~Navab.
\newblock {SSD-6D: Making RGB-based 3D detection and 6D pose estimation great
  again}.
\newblock In {\em Proceedings of the International Conference on Computer
  Vision (ICCV 2017), Venice, Italy}, pages 22--29, 2017.

\bibitem{kendall2017geometric}
A.~Kendall, R.~Cipolla, et~al.
\newblock Geometric loss functions for camera pose regression with deep
  learning.
\newblock In {\em Proc. CVPR}, volume~3, page~8, 2017.

\bibitem{kendall2015posenet}
A.~Kendall, M.~Grimes, and R.~Cipolla.
\newblock {PoseNet: A convolutional network for real-time 6-DOF camera
  relocalization}.
\newblock In {\em Proceedings of the IEEE international conference on computer
  vision}, pages 2938--2946, 2015.

\bibitem{kuznietsov2017semi}
Y.~Kuznietsov, J.~St{\"u}ckler, and B.~Leibe.
\newblock Semi-supervised deep learning for monocular depth map prediction.
\newblock In {\em Proc. of the IEEE Conference on Computer Vision and Pattern
  Recognition}, pages 6647--6655, 2017.

\bibitem{laina2016deeper}
I.~Laina, C.~Rupprecht, V.~Belagiannis, F.~Tombari, and N.~Navab.
\newblock Deeper depth prediction with fully convolutional residual networks.
\newblock In {\em 3D Vision (3DV), 2016 Fourth International Conference on},
  pages 239--248. IEEE, 2016.

\bibitem{li2018deep}
R.~Li, K.~Xian, C.~Shen, Z.~Cao, H.~Lu, and L.~Hang.
\newblock Deep attention-based classification network for robust depth
  prediction.
\newblock {\em arXiv preprint arXiv:1807.03959}, 2018.

\bibitem{MegaDepthLi18}
Z.~Li and N.~Snavely.
\newblock Megadepth: Learning single-view depth prediction from internet
  photos.
\newblock In {\em Computer Vision and Pattern Recognition (CVPR)}, 2018.

\bibitem{liu2015deep}
F.~Liu, C.~Shen, and G.~Lin.
\newblock Deep convolutional neural fields for depth estimation from a single
  image.
\newblock In {\em Proceedings of the IEEE Conference on Computer Vision and
  Pattern Recognition}, pages 5162--5170, 2015.

\bibitem{liu2018intriguing}
R.~Liu, J.~Lehman, P.~Molino, F.~P. Such, E.~Frank, A.~Sergeev, and
  J.~Yosinski.
\newblock An intriguing failing of convolutional neural networks and the
  coordconv solution.
\newblock {\em arXiv preprint arXiv:1807.03247}, 2018.

\bibitem{mayer2016large}
N.~Mayer, E.~Ilg, P.~Hausser, P.~Fischer, D.~Cremers, A.~Dosovitskiy, and
  T.~Brox.
\newblock A large dataset to train convolutional networks for disparity,
  optical flow, and scene flow estimation.
\newblock In {\em Proceedings of the IEEE Conference on Computer Vision and
  Pattern Recognition}, pages 4040--4048, 2016.

\bibitem{ronneberger2015u}
O.~Ronneberger, P.~Fischer, and T.~Brox.
\newblock {U-net: Convolutional networks for biomedical image segmentation}.
\newblock In {\em International Conference on Medical image computing and
  computer-assisted intervention}, pages 234--241. Springer, 2015.

\bibitem{saxena2009make3d}
A.~Saxena, M.~Sun, and A.~Y. Ng.
\newblock {Make3D: Learning 3D scene structure from a single still image}.
\newblock {\em IEEE transactions on pattern analysis and machine intelligence},
  31(5):824--840, 2009.

\bibitem{schonberger2016structure}
J.~L. Schonberger and J.-M. Frahm.
\newblock Structure-from-motion revisited.
\newblock In {\em Proceedings of the IEEE Conference on Computer Vision and
  Pattern Recognition}, pages 4104--4113, 2016.

\bibitem{shrivastava2017learning}
A.~Shrivastava, T.~Pfister, O.~Tuzel, J.~Susskind, W.~Wang, and R.~Webb.
\newblock {Learning from Simulated and Unsupervised Images through Adversarial
  Training.}
\newblock In {\em CVPR}, pages 2242--2251, 2017.

\bibitem{silberman2012indoor}
N.~Silberman, D.~Hoiem, P.~Kohli, and R.~Fergus.
\newblock Indoor segmentation and support inference from rgbd images.
\newblock In {\em European Conference on Computer Vision}, pages 746--760.
  Springer, 2012.

\bibitem{sundermeyer2018implicit}
M.~Sundermeyer, Z.-C. Marton, M.~Durner, M.~Brucker, and R.~Triebel.
\newblock {Implicit 3D Orientation Learning for 6D Object Detection from RGB
  Images}.
\newblock In {\em Proceedings of the European Conference on Computer Vision
  (ECCV)}, pages 699--715, 2018.

\bibitem{tang2018ba}
C.~Tang and P.~Tan.
\newblock Ba-net: Dense bundle adjustment network.
\newblock {\em arXiv preprint arXiv:1806.04807}, 2018.

\bibitem{tatarchenko2017octree}
M.~Tatarchenko, A.~Dosovitskiy, and T.~Brox.
\newblock Octree generating networks: Efficient convolutional architectures for
  high-resolution 3d outputs.
\newblock In {\em Proc. of the IEEE International Conf. on Computer Vision
  (ICCV)}, volume~2, page~8, 2017.

\bibitem{tateno2017cnn}
K.~Tateno, F.~Tombari, I.~Laina, and N.~Navab.
\newblock {CNN-SLAM: Real-time dense monocular SLAM with learned depth
  prediction}.
\newblock In {\em Proceedings of the IEEE Conference on Computer Vision and
  Pattern Recognition (CVPR)}, volume~2, 2017.

\bibitem{torralba2011unbiased}
A.~Torralba and A.~A. Efros.
\newblock Unbiased look at dataset bias.
\newblock In {\em Computer Vision and Pattern Recognition (CVPR), 2011 IEEE
  Conference on}, pages 1521--1528. IEEE, 2011.

\bibitem{Uhrig2017THREEDV}
J.~Uhrig, N.~Schneider, L.~Schneider, U.~Franke, T.~Brox, and A.~Geiger.
\newblock {Sparsity Invariant CNNs}.
\newblock In {\em International Conference on 3D Vision (3DV)}, 2017.

\bibitem{ummenhofer2017demon}
B.~Ummenhofer, H.~Zhou, J.~Uhrig, N.~Mayer, E.~Ilg, A.~Dosovitskiy, and
  T.~Brox.
\newblock {DeMoN: Depth and Motion Network for learning monocular stereo}.
\newblock In {\em IEEE Conference on computer vision and pattern recognition
  (CVPR)}, volume~5, page~6, 2017.

\bibitem{wang2015towards}
P.~Wang, X.~Shen, Z.~Lin, S.~Cohen, B.~Price, and A.~L. Yuille.
\newblock Towards unified depth and semantic prediction from a single image.
\newblock In {\em Proceedings of the IEEE Conference on Computer Vision and
  Pattern Recognition}, pages 2800--2809, 2015.

\bibitem{wang2017deepvo}
S.~Wang, R.~Clark, H.~Wen, and N.~Trigoni.
\newblock {DeepVO: Towards end-to-end visual odometry with deep recurrent
  convolutional neural networks}.
\newblock In {\em Robotics and Automation (ICRA), 2017 IEEE International
  Conference on}, pages 2043--2050. IEEE, 2017.

\bibitem{wang2018end}
S.~Wang, R.~Clark, H.~Wen, and N.~Trigoni.
\newblock End-to-end, sequence-to-sequence probabilistic visual odometry
  through deep neural networks.
\newblock {\em The International Journal of Robotics Research},
  37(4-5):513--542, 2018.

\bibitem{weerasekera2018just}
C.~S. Weerasekera, T.~Dharmasiri, R.~Garg, T.~Drummond, and I.~Reid.
\newblock Just-in-time reconstruction: Inpainting sparse maps using single view
  depth predictors as priors.
\newblock {\em arXiv preprint arXiv:1805.04239}, 2018.

\bibitem{weyand2016planet}
T.~Weyand, I.~Kostrikov, and J.~Philbin.
\newblock Planet-photo geolocation with convolutional neural networks.
\newblock In {\em European Conference on Computer Vision}, pages 37--55.
  Springer, 2016.

\bibitem{xiao2013sun3d}
J.~Xiao, A.~Owens, and A.~Torralba.
\newblock {SUN3D: A database of big spaces reconstructed using SfM and object
  labels}.
\newblock In {\em Proceedings of the IEEE International Conference on Computer
  Vision}, pages 1625--1632, 2013.

\bibitem{zhang2018deep}
Y.~Zhang and T.~Funkhouser.
\newblock {Deep Depth Completion of a Single RGB-D Image}.
\newblock In {\em Proceedings of the IEEE Conference on Computer Vision and
  Pattern Recognition}, pages 175--185, 2018.

\bibitem{zhou2018deeptam}
H.~Zhou, B.~Ummenhofer, and T.~Brox.
\newblock {DeepTAM}: Deep tracking and mapping.
\newblock In {\em European Conference on Computer Vision (ECCV)}, 2018.

\bibitem{zhou2017unsupervised}
T.~Zhou, M.~Brown, N.~Snavely, and D.~G. Lowe.
\newblock Unsupervised learning of depth and ego-motion from video.
\newblock In {\em CVPR}, pages 6612--6619, 2017.

\bibitem{zhou2016view}
T.~Zhou, S.~Tulsiani, W.~Sun, J.~Malik, and A.~A. Efros.
\newblock View synthesis by appearance flow.
\newblock In {\em European conference on computer vision}, pages 286--301.
  Springer, 2016.

\bibitem{zoran2015learning}
D.~Zoran, P.~Isola, D.~Krishnan, and W.~T. Freeman.
\newblock Learning {Ordinal} {Relationships} for {Mid}-{Level} {Vision}.
\newblock In {\em 2015 {IEEE} {International} {Conference} on {Computer}
  {Vision} ({ICCV})}, pages 388--396, Dec. 2015.

\end{thebibliography}
}

\end{document}